# Design of scalable orthogonal digital encoding architecture for large-area flexible tactile sensing in robotics

Weijie Liu, Ziyi Qiu, Shihang Wang, Deqing Mei and Yancheng Wang*, *Senior Member, IEEE*

*Abstract*—Human-like embodied tactile perception is crucial for the next-generation intelligent robotics. Achieving large-area, full-body soft coverage with high sensitivity and rapid response, akin to human skin, remains a formidable challenge due to critical bottlenecks in encoding efficiency and wiring complexity in existing flexible tactile sensors, thus significantly hinder the scalability and real-time performance required for human skin-level tactile perception. Herein, we present a new architecture employing code division multiple access- inspired orthogonal digital encoding to overcome these challenges. Our decentralized encoding strategy transforms conventional serial signal transmission by enabling parallel superposition of energy-orthogonal base codes from distributed sensing nodes, drastically reducing wiring requirements and increasing data throughput. We implemented and validated this strategy with off-the-shelf 16-node sensing array to reconstruct the pressure distribution, achieving a temporal resolution of 12.8 ms using only a single transmission wire. Crucially, the architecture can maintain sub-20ms latency across orders-of-magnitude variations in node number (to thousands of nodes). By fundamentally redefining signal encoding paradigms in soft electronics, this work opens new frontiers in developing scalable embodied intelligent systems with human-like sensory capabilities.

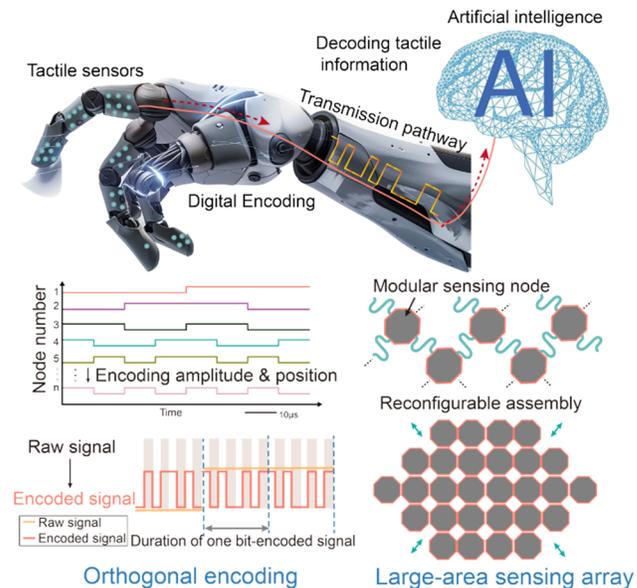

**Fig. 1. Pipeline overview**: Orthogonal encoding-based method and modular assembly for large-area tactile sensing in robotics.

## I. INTRODUCTION

Tactile perception constitutes a cornerstone of biological intelligence, enabling humans to perform delicate object manipulation, environmental mapping, and adaptive motor control through spatially distributed mechanoreceptors [1]. Artificial tactile perception seeks to mirror these capabilities in robotic systems, a pursuit that is increasingly critical for next-generation robots to autonomously explore, make decisions, and learn in unstructured environments [2]. Benefiting from advances in functional materials and devices, flexible tactile sensors and electronic skin that mimic the properties and functions of human skin have made significant progress [3, 4]. Similar to how human soft skin envelops rigid bones, covering the surface of robots with soft electronic skin not only enables the detection of multidimensional tactile information through direct physical contact but also increases the contact area and provides a degree of protection to the robot body [5]. Moreover, with 2D-to-3D design strategies [6, 7], flexible electronic skin can adapt to changes in the robot's structural form without the need for redesigning to fit specific robotics.

Although major strides have been made in flexible tactile sensor and electronic skin technologies, most implementations are still limited to small-scale tactile perception in localized areas, such as robotic hands [8]. Replicating large-area, full-body soft coverage like human skin while maintaining high sensitivity and rapid response remains an unresolved challenge. Excessive wiring requirements, slow encoding efficiency, and difficulties maintaining flexibility at scale continue to hinder progress. The core issue stems from the von Neumann-style readout and encoding paradigm pervasive in tactile arrays, which extends sensing from a single point to surface and generate frames of tactile signals [9]. Conventional N×N matrix scanning requires O(N) wires and exhibits $O(N^2)$ time complexity-a fundamental scaling law causing exponential performance degradation with array size, rendering scaling-up both cumbersome and costly [10]. When the sensing array needs to adapt to various robotic platforms-differing in size or curvature, this rigid architecture often necessitates a complete redesign, undercutting its broader applicability.

To achieve large-area flexible tactile sensing, a scalable architecture with minimal wiring and fast response is critical.

Research supported by the National Natural Science Foundation of China (52475572 and 52175522), Zhejiang Province Natural Science Foundation of China (LZ24E050003), and Key Research and Development Program of Zhejiang Province (2025C0003).

Y.C. Wang, D.Q. Mei are with State Key Laboratory of Fluid Power & Mechatronic Systems, School of Mechanical Engineering, Zhejiang University, Hangzhou, 310058, China (Corresponding author: Yancheng Wang, phone: 86-13675828104; e-mail: yanchwang@zju.edu.cn).

W.J. Liu, Z.Y. Qiu, S.H. Wang are with Zhejiang Key Laboratory of Advanced Equipment Manufacturing and Measurement Technology, School of Mechanical Engineering, Zhejiang University, Hangzhou, 310058, China.

Nilsson [11] proposed a single-wire acquisition based on filter characteristics and shimojo et al. [12] proposed grid structures leveraging current density. The inability to decode both pressure magnitude and position, coupled with poor scalability, significantly limits their applicability. Cheng et al. [13, 14] developed extended architectures using serial buses(UART or I²C). However, it requires a complex arbitration mechanism and is challenging to integrate with flexible devices. Emerging architectures employing frequency-domain encoding schemes [15], bio-inspired asynchronous encoding [16], and position-encoded spike spectrum [17] partially address wiring complexity and speed. But inaccuracies are exacerbated in order to achieve minimal wiring and rapid encoding, and much valuable tactile time-domain information is often sacrificed. Achieving high-precision encoding and decoding of both tactile distribution and intensity, while maintaining simple deployment and compatibility with flexible sensing elements, remains a significant challenge.

In this study, we proposed a new scalable architecture employing code division multiple access- inspired orthogonal digital encoding for large-area flexible tactile sensing in robotics, as shown in Fig. 1. The tactile signals are encoded based on a set of energy-orthogonal base codes by decentralized intelligence, enabling the transmission of the encoded signals from all the sensing nodes in parallel in a superposition, thus drastically reducing wiring requirements and increasing data throughput without loss of temporal information. We deployed and validated this architecture with off-the-shelf 4×4 sensing array. The experimental results show that the encoding method is capable of achieving fast encoding and transmission of tactile signals with a single wire of 12.8 ms temporal resolution, and that the performance is highly tunable for an arbitrary number of sensing nodes, and that an almost constant tactile signal latency can be achieved for a fairly large range of the number of sensing nodes, underscoring its scalability for real-time, large-area tactile sensing in next-generation robotic systems.

## II. Design of Tactile encoding architecture

### A. Scalable Orthogonal Digital Encoding Architecture

To enable large-area scalability, minimal wiring, and rapid tactile sensing, we propose a scalable orthogonal digital encoding architecture for flexible robotic tactile skin, drawing inspiration from next-generation multiple access technologies. As shown in Fig. 2, the tactile sensing system comprises distributed sensing nodes as fundamental units. Each node contains a sensitive element that converts mechanical stimuli into electrical signals, generating raw tactile signals represented as 10-12 bit binary digital data (with bit-length denoted as $k$, determined by analog-to-digital converter resolution). Subsequently the raw tactile signals are encoded, and the encoding processes of different sensing nodes are independent of each other. For a total of $n$ sensing nodes, each is assigned a unique $n$-dimensional encoding vector $\mathbf{C}_i$, which is defined as

$$\mathbf{C}_i = [c_{i1}, c_{i2}, c_{i3} \ldots c_{in}], \quad i=1,2,\ldots,n \quad (1)$$

where $c_{i1}, c_{i2} \ldots c_{in}$ is the $n$ components of the encoding vector $\mathbf{C}_i$. At the same time, the $n$-dimensional encoding vectors of different nodes are orthogonal to each other. ensuring uniqueness across nodes

$$\mathbf{C}_i \mathbf{C}_j = \sum_{k=1}^{n} c_{ik} c_{jk} = 0, \quad \forall i \neq j \quad (2)$$

This encoding vector forms the basis for signal encoding. During the encoding process, each bit of the $k$-bit raw tactile signal is sequentially encoded and transmitted. If a given bit is 1, the encoding vector itself is used; if the bit is 0, the inverse of the encoding vector (each component is the opposite of the original) is used. Once all $k$ bits are encoded, the process for that sensing node is complete. Consequently, the generation of robust, transmittable analog encoded signals relies on $n$-dimensional orthogonal vectors. To facilitate this, we adopt bi-valued orthogonal encoding vectors, where each component $c_{i1}, c_{i2} \ldots c_{in}$ can take one of two distinct values, enabling their representation through discrete high and low voltage levels.

There are two common types of bi-valued orthogonal vector sets, one is $n$-dimensional orthogonal basis vectors containing only 0 and 1 elements, and the other is interleaved orthogonal vectors containing only 1 and -1 elements. As shown in Fig. 3(a), the two distinct values of a bi-valued orthogonal vector can be represented as high and low voltage levels. For example, if the vector elements are 0 and 1, the high level corresponds to 1 and the low level to 0. If the elements are 1 and −1, the high level represents 1 and the low level represents −1. Building on this concept, each component of the encoding vector can be represented by a sustained voltage level, while changes in component values can be realized through high-to-low or low-to-high voltage transitions. Given an orthogonal vector of length $n$, this can be simulated through a sequence of $n$ voltage level, and each hold for a duration $T$. Thus, simulating an orthogonal vector requires $n \times T$ time, while encoding a $k$-bit raw tactile signal requires $k \times n \times T$ time. As shown in Fig. 3(b), one of the nodes in eight-node sensing network encodes two bits of binary data of a raw tactile signal during the encoding process.

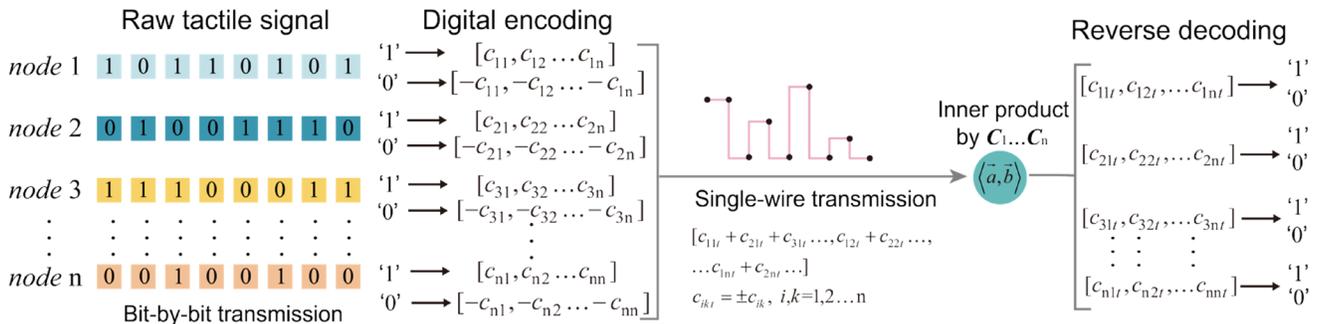

**Fig. 2. Schematic diagram of the scalable orthogonal digital encoding architecture.** The signals of distributed tactile nodes are encoded based on a set of energy-orthogonal base signals, enabling the transmission of the encoded signals from all the sensing nodes in parallel in a superposition.

Due to the orthogonality between encoding vectors, the encoded signals from all nodes can be superimposed and transmitted simultaneously in the time domain. The original signals can then be reconstructed through orthogonal decoding, allowing the acquisition of tactile signals from multiple sensing nodes using a single data wiring. The orthogonal decoding can be defined as

$$\mathbf{S}_t \mathbf{C}_i = \sum_{k=1}^{n} \mathbf{C}_{k_t} \mathbf{C}_i = \begin{cases} > 0, \text{ encoded bit=1} \\ < 0, \text{ encoded bit=0} \\ = 0, \text{ no encoded bit} \end{cases} \quad (3)$$

$$\mathbf{C}_{k_t} = \pm \mathbf{C}_k, \ i = 1, 2, \ldots n$$

where $\mathbf{S}_t$ is the superimposed signal acquired at moment $t$.

For large-area scalability, the architecture must accommodate an arbitrary number of sensing nodes. To this end, efficiently generating $n$ pairs of $n$-dimensional bi-valued orthogonal vectors is essential. Hadamard matrices provide a convenient solution, offering bi-valued orthogonal vectors of $2^n$ order with elements $\{\pm1\}$, which can be conveniently generated through iterative programming. Each row of the Hadamard matrix is mutually orthogonal, making it suitable for use as the encoding vector of sensing nodes. When the number of sensing nodes does not match a power of 2, the smallest greater power of 2 can be used. For instance, a 500-node sensing array can utilize a 512-order Hadamard matrix to generate 500 orthogonal vectors of 512 dimensions, while an 800-node array can employ a 1024-order Hadamard matrix to generate 800 orthogonal vectors of 1024 dimensions. Theoretically, the proposed architecture is universally scalable and can be applied to any number of sensing nodes, enabling seamless expansion for large-area applications. Moreover, since each bit of the raw tactile signal is sequentially encoded and transmitted, the tactile time-domain information is preserved throughout the encoding process and can be losslessly decoded.

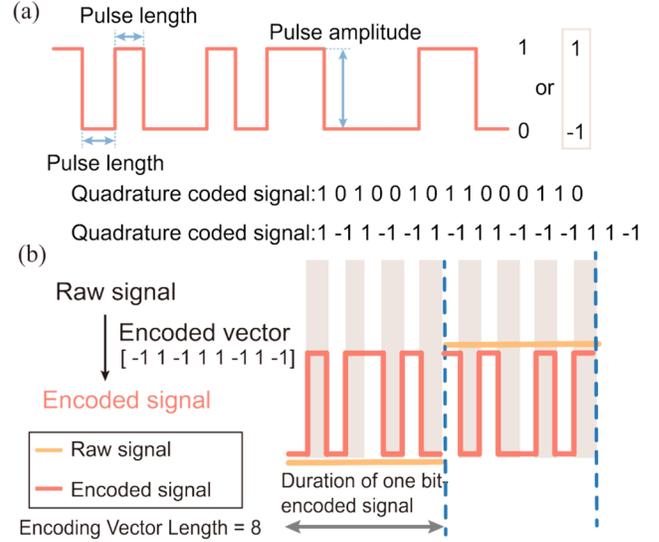

**Fig. 3.** (a) Schematic diagram to generate transmittable analog encoded signals relying on bi-valued orthogonal vectors. (b) an example that one of the nodes in eight-node sensing network encodes two bits of binary data during the encoding process

There are two main characteristic parameters of the encoded signal, which are the pulse length (the level holding time corresponding to one of the smallest components of the encoded vector) and the pulse amplitude (the difference between the high and low levels). The former mainly affects the response time of the encoding architecture, and the latter mainly affects the maximum number of simultaneous decoding. A major advantage of the proposed architecture is its adaptability. As mentioned earlier, the time to transmit a $k$-bit raw tactile signal is $k \times n \times T$. We can adjust $T$ according to the number of sensing nodes as needed to achieve continuously tunable response time over a wide range, and can even achieve a constant transmission speed for any number of sensing nodes in the tactile sensing system.

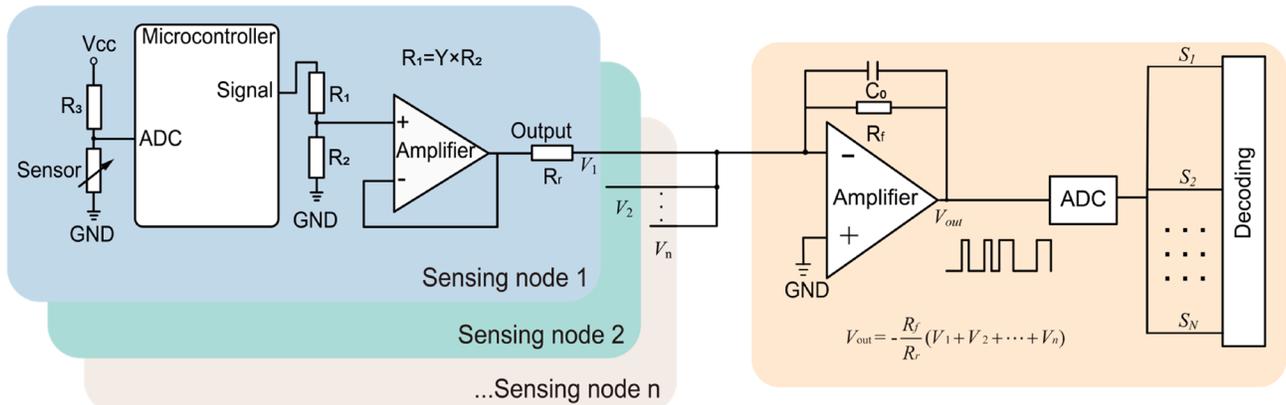

**Fig. 4. Hardware circuit design for tactile signal encoding.** Each sensing node comprises a sensing unit and an encoding circuit. An inverting summing amplifier with a unity gain configuration is used to superimposed all encoded signals.

### B. Hardware Design for Orthogonal Digital Encoding

To implement tactile signal encoding in hardware, we designed the sensing node circuit and proposed a corresponding encoded signal generation method. As shown in Fig. 4, each sensing node comprises a sensing unit (pressure-sensitive element) and an encoding circuit. In this paper, as a demonstration, the sensing units are implemented using piezoresistive elements. However, in principle, any sensing mechanism capable of generating signals compatible with AD acquisition can be used. The pressure sensitive element is connected to a voltage divider circuit, which measures the voltage and converts it into a 10-bit digital raw tactile signal using a microcontroller. Each microcontroller is internally assigned an $n$-dimensional orthogonal encoding vector, corresponding to a predefined encoding scheme. The components of encoding vector are represented through successive high- and low-level transitions of the microcontroller's IO ports. When the resistance of the sensing unit exceeds a predefined threshold, the microcontroller generates and outputs the encoded signal based on the assigned encoding scheme. The encoded signals from the sensing nodes are first passed through a voltage divider circuit(Y-fold reduction) to proportionally attenuate the amplitude so that it does not exceed the common-mode input voltage of the subsequent amplifiers. The signals are then buffered using a voltage follower to minimize interference between signals from different nodes. Following attenuation, each node's encoded signal is fed into an inverting summing amplifier, where the signals are superimposed with a unity gain configuration. The aggregated signal is then digitized via AD acquisition and transmitted to a host computer for further decoding. The encoding circuit is designed with both integration and performance in mind. All components can be selected as ultra-miniature packages to achieve maximum functionality within a minimal footprint. This compact design facilitates future development into a decentralized, modular sensing unit structure, enhancing scalability and flexibility in large-area tactile sensing applications.

## III. EXPERIMENTAL SETUP AND PROCEDURE

### A. Tactile Sensing Array and Characterization

To validate the feasibility of the proposed encoded architecture, a multi-node tactile sensing array must first be developed. While modular sensing nodes offer a scalable solution for large-area tactile skin, we opted to fabricate a 16-node flexible tactile sensing array to simplify the design and manufacturing process. As shown in Fig. 5, the array consists of four functional layers (from top to bottom): an encapsulation layer, a piezoresistive sensing layer, an interdigital electrode layer, and a PI substrate layer. The piezoresistive sensing layer is composed of commercially available conductive rubber (purchased from INABA, Japan), which exhibits a significant decrease in resistance upon the application of pressure. The resistance changes in this layer are detected by a planar interdigital electrode, patterned using photolithography and etching processes. Each individual sensing unit measures 5.5 mm × 5.5 mm with a thickness of 1 mm. For pressure characterization, the tactile sensing array was placed on a custom calibration platform. The motion of the loading bar was precisely controlled by a linear stage (Newport UTS series, USA), which moved vertically to apply pressure to the sensor. The applied force was measured using an ATI 6-axis force sensor (Nano 17, ATI Industrial Automation), offering a force resolution of 0.01 N.

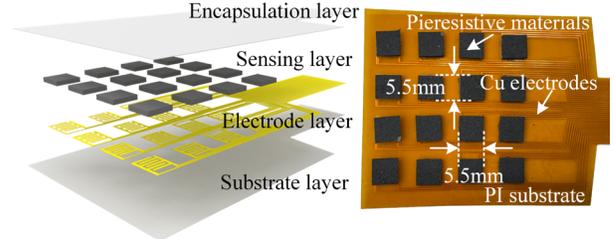

Fig. 5. Structural schematic diagram(left) and developed prototype(right) of the 16-node sensing array.

### B. Tactile Signal Encoding and Decoding

Building on the previously designed encoding hardware, a single circuit board was used to encode tactile signals from the 16-node sensing array. The encoded single-channel output signal was acquired using an oscilloscope(MSOX2024A, Keysight) and transmitted to a computer for processing and decoding in MATLAB. The evaluation process primarily focused on analyzing the temporal resolution and error of the encoded signal, while the decoded results were used to reconstruct the pressure distribution across the array.

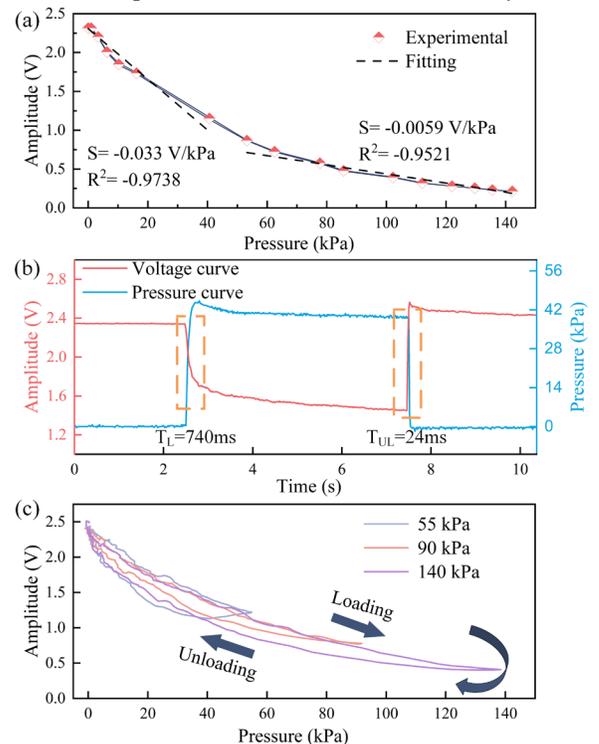

Fig. 6. **Sensing performance of the sensing array**. (a) pressure sensing characterization of sensitivity and range. (b) the response time and (c) hysteresis of the sensing array.

## IV. RESULTS AND EVALUATIONS

### A. Characterization Results of the tactile sensing array

The pressure sensing performance of the tactile sensing unit was investigated, as shown in Fig. 6(a)–(c). The relationship between applied pressure and the corresponding voltage amplitude change was first studied. As shown in Fig. 6(a), the sensor voltage value gradually decreases with the pressure between the loading bar and the sensor increasing. It can be seen clearly that the developed tactile sensing unit generally exhibits two-stage stable linear sensitivity: -0.033V/kPa when applied normal force from 0 to 50 kPa, and -0.0059V/kPa from 50 to 140 kPa. The decline in sensitivity of the second stage can be attributed to the saturation of deformation in the sensing materials during this phase. When the pressure increases further, the change in voltage amplitude is small, so the sensing range of the sensor is 0~140 kPa.

Before further applications, the response time of the pressure-sensitive element was also characterized. A normal force of 40 kPa was applied to the sensing unit at a rate of 1 Hz and held for 5 s. The results demonstrated that the fabricated sensor exhibited a response time of 740 ms and a recovery time of 24 ms, as shown in Fig. 6(b). Hysteresis is an important parameter for flexible tactile sensors, which was tested by applying different normal forces to the tactile sensor with a 0.1 Hz of the motion platform. In Fig.6(c), the tactile sensor's hysteresis was 33.87%, 10.99%, and 10.62% when the applied normal forces were 55, 90, and 140 kPa, respectively. The hysteresis may be attributed to the viscoelasticity of the sensing materials during unloading.

### B. Encoding Signal Generation and Evaluation

To evaluate the performance of the encoded signals, we set the minimum pulse duration for each encoded vector component to 50 μs. According to the encoding principle, successive 16 × 10 × 50 μs segments encoded a frame of 10-bit binary raw tactile signal. Fig. 7(a) presents the encoded signal frames generated by a sensing node in response to applied pressure. The resulting waveform resembles a square wave with an amplitude of −300 mV and a duration of 8 ms. Due to hardware timing variations, the actual measured duration of the encoded frames fluctuates between 8 ms and 8.8 ms. To ensure clear separation between consecutive frames, a temporal gap must be introduced so that there is enough redundant space between two frames of encoded signal to be accurately decoded. This gap defines the temporal resolution of the encoding architecture. We tested a temporal resolution of 12.8 ms, as shown in Fig. 7(a), where a complete tactile encoding frame is generated every 12.8 ms.

We also observed the minimum pulse-level transition corresponding to individual components of the encoded vector, as shown in Fig. 7(b). The level duration accuracy can reach within 10%, with level switching time as fast as 0.1 to 0.2 μs, rendering them practically negligible. The pulse amplitude becomes 300 mV after attenuation by a voltage divider circuit to improve the recognizability of the pulse waveform and to avoid saturation of the output of the inverting summing amplifier. Notably, if multiple consecutive components of the encoding vector share the same value, the resulting waveform exhibits a continuous level, as shown in Fig. 7(c). The measured duration of two continuous-level waveforms was 104.3 μs, demonstrating that the duration accuracy remains within 10%, without accumulating additional error over multiple cycles.

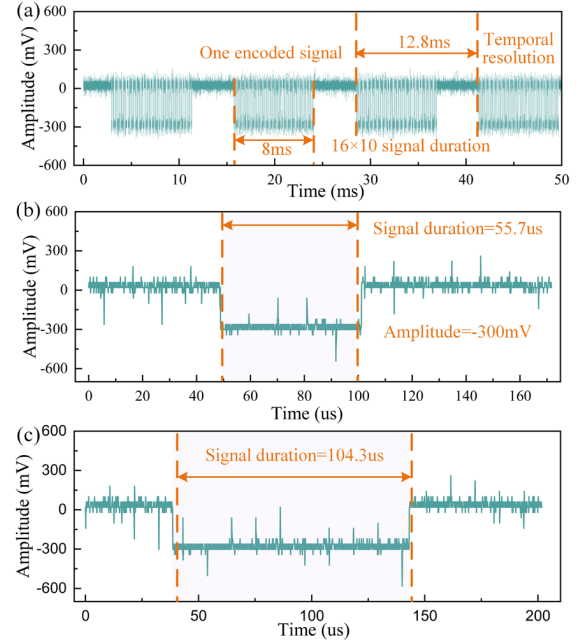

Fig. 7. **Encoding signal evaluation** (a) diagram of encoded signal frames and temporal resolution (b) minimum pulse level duration corresponding to one of the encoding vector components. (c) continuous minimum pulse level duration corresponding to two of the encoding vector components.

### C. Decoding results of tactile signal

Accurately decoding the pressure distribution and amplitude of the sensing array is crucial for validating the effectiveness of the proposed encoding architecture. To this end, we conducted single-point and multi-point pressure decoding experiments, as shown in Fig. 8(a)-(b).

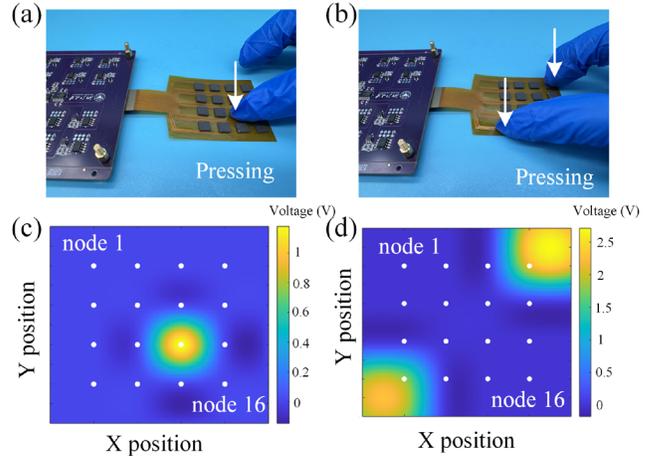

Fig. 8. **The single-point and multi-point pressure experiments and decoding results** (a)-(b) Experimental diagrams of single-point and multi-point pressing (c)-(d) the decoding results of the tactile signal.

The single-channel output signal of the sensing array was acquired and processed frame by frame using orthogonal decoding. The energy-orthogonal patterns of different nodes enable the separation and extraction of each node's raw tactile signal from the superimposed signal. As illustrated in Fig. 8(c)-(d), the decoding process successfully extracted the binary sequence 0101101110 (corresponding to a voltage of 1.18V) from node 11 under a single-point conditions, and the binary sequences 1000110000 (corresponding to a voltage of 1.80 V) from node 4 and 1010011001 (corresponding to a voltage of 2.14 V) from node 13 under multi-point pressure conditions. The results demonstrate that there is minimal crosstalk between different nodes in the sensing array, confirming the effectiveness and robustness of the proposed approach. To reduce power consumption in the proposed architecture, we introduced the event-driven mechanism. For each sensing node, if the voltage variation remains below a predefined threshold, the node remains inactive during that encoding frame and does not output a signal.

To further demonstrate the feasibility of our method for large-area robotic flexible tactile skin, we developed a scalable modular flexible sensing node, as shown in Fig. 9. Consistent with our previous design, each node comprises a sensing element and an encoding circuit. The nodes are mechanically and electrically interconnected via stretchable electrodes and share a common output channel. Currently the smallest modular units can be up to 1 cm$^2$ in size, and the integration of multiple sensitive units on a single unit can further increase the spatial resolution. This modular design allows the sensing nodes to be assembled like LEGO bricks, enabling the construction of large-area sensing arrays with arbitrary scalability, making it highly adaptable for various robotic applications. In general, our work opens new frontiers in developing scalable embodied intelligent systems with human-like sensory capabilities.

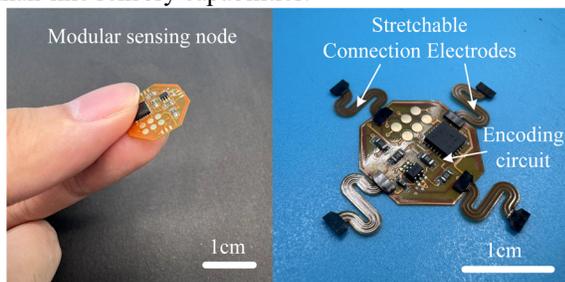

**Fig. 9. Modular sensing node based on scalable orthogonal digital encoding architecture**.

## V. Conclusions

In this paper, we proposed a scalable architecture employing code division multiple access- inspired orthogonal digital encoding for tactile sensing. The tactile signals are encoded based on a set of energy-orthogonal base codes by decentralized intelligence, enabling the transmission of the encoded signals from all the sensing nodes in parallel in a superposition, thus drastically reducing wiring requirements and increasing data throughput. We deployed and validated this architecture with off-the-shelf 4×4 sensing array. The experimental results show that the encoding method is capable of achieving fast encoding and transmission of tactile signals with a single wire of 12.8 ms temporal resolution, and that the performance is highly tunable for an arbitrary number of sensing nodes, and that an almost constant tactile signal latency can be achieved for a fairly large range of the number of sensing nodes, underscoring its scalability for real-time, large-area tactile sensing in next-generation robotic systems.